\documentclass[letterpaper, 10 pt, conference]{ieeeconf}  % Comment this line out
\IEEEoverridecommandlockouts                              % This command is only
\usepackage{graphics} % for pdf, bitmapped graphics files
\usepackage{epsfig} % for postscript graphics files
\usepackage{mathptmx} % assumes new font selection scheme installed
\usepackage{times} % assumes new font selection scheme installed
\usepackage{amsmath} % assumes amsmath package installed
\usepackage{amssymb}  % assumes amsmath package installed
\usepackage{graphicx}
\usepackage{tabto}
\usepackage{hyperref}
\usepackage{amsmath}
\usepackage{makecell}
\usepackage{caption}
\usepackage{subcaption}
\usepackage{algorithmic}
\usepackage[linesnumbered,ruled]{algorithm2e}

\title{\LARGE \bf
Kinematics \& Dynamics Library for Baxter Arm}

\author{Akshay Kumar, Ashwin Sahasrabudhe, Chaitanya Perugu, Sanjuksha Nirgude, Aakash Murugan
\thanks{All the authors are with the Department of Robotics Engineering, Worcester Polytechnic Institute, USA}
}

\begin{document}

\maketitle
\thispagestyle{empty}
\pagestyle{empty}

%%%%%%%%%%%%%%%%%%%%%%%%%%%%%%%%%%%%%%%%%%%%%%%%%%%%%%%%%%%%%%%%%%%%%%%%%%%%%%%%
\begin{abstract}

The Baxter robot is a standard research platform used widely in research tasks, supported with an SDK provided by the developers, Rethink Robotics. Despite the ubiquitous use of the robot, the official software support is sub-standard. Especially, the native IK service has a low success rate and is often inconsistent. This unreliable behavior makes Baxter difficult to use for experiments and the research community is in need of a more reliable software support to control the robot. We present our work towards creating a Python based software library supporting the kinematics and dynamics of the Baxter robot. Our toolbox contains implementation of pose and velocity kinematics with support for Jacobian operations for redundancy resolution. We present the implementation and performance of our library, along with a comparison with PyKDL.

\textbf{\textit{ Keywords---} Baxter Research Robot, Manipulator Kinematics, Iterative IK, Dynamical Model, Redundant Manipulator}

\end{abstract}

%%%%%%%%%%%%%%%%%%%%%%%%%%%%%%%%%%%%%%%%%%%%%%%%%%%%%%%%%%%%%%%%%%%%%%%%%%%%%%%%
\section{Introduction}

\subsection{Introduction to Baxter}

Baxter is a dual arm humanoid robot developed by Rethink Robotics Inc., with 7 degrees of freedom on each arm manipulator, hence, falling under the \textit{kinematically redundant} category. Baxter's arms are actuated by Serial Elastic Actuators (SEAs) which provide inherent compliance to the arm for safety purposes. Baxter has three offsets in its arm kinematic structure. Each 7-DOF arm has a 2-DOF offset shoulder, a 2-DOF elbow and a 3-DOF offset wrist. Both arms include angle position and joint torque sensing. Due to the offsets, no three consecutive frames meet at a common origin, hence, according to Pieper's principle, there is no analytical solution for 7-DOF Inverse Pose Kinematics (IPK) of Baxter's arm manipulator \cite{c8}. 

\begin{figure}[h!]
\centering
\includegraphics[width=6cm]{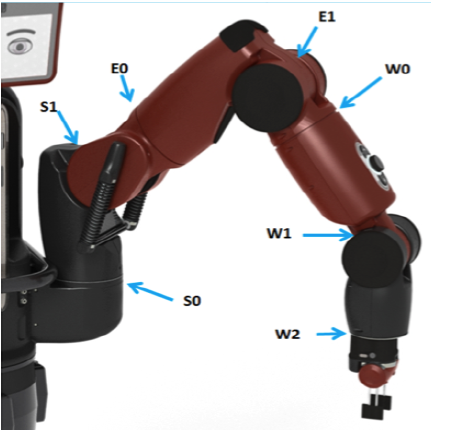}
\caption{Baxter's 7-DOF arm}
\label{fig:joints} 
\end{figure}

In Fig. \ref{fig:joints} the shoulder joints are represented by S0 (shoulder roll) and S1 (shoulder pitch), elbow joints by E0 (elbow roll) and E1 (elbow pitch), and wrist joints by W0 (wrist roll), W1 (wrist pitch) and W2 (wrist roll). 

Planning a robot's motion requires understanding of the relationship between the actuators we can control and the robot's resulting position in the environment. \textit{Forward Kinematics} of a robotic arm is the process used to find the position of the end-effector of the robot using the knowledge of the angle of each joint. If we need to find the angle of each joint for a particular end-effector position, we need to invert this relationship. This process is known as \textit{Inverse Kinematics}. \textit{Dynamics} of a robot provides the relationship between actuations and contact forces, acceleration and the resulting motion trajectories. Through this project, we make this process easier for the user of the Baxter robot by developing a Python library to perform the forward and inverse pose kinematics and dynamics of a well-known humanoid robot, Baxter.

\subsection{Kinematics}

Robot mechanisms consist of number of rigid bodies connected by joints. The position and orientation of these rigid bodies in space are together termed as a \textit{pose}. The robot kinematics describes the pose, velocity and acceleration of the rigid bodies that make up their robot mechanism. A \textit{kinematic joint} is a connection between two bodies that allows for relative motion. Kinematics of the robot consists of two processes, forward pose kinematics and inverse pose kinematics. In forward pose kinematics of a serial manipulator, we find the position and orientation of the end-effector frame of interest from the joint values. In the process of inverse kinematics, we find the values of joint positions that result in the given values of position and orientation of the end-effector relative to the base frame.  

\subsection{Dynamics}
Dynamics of a robot describes the relationship between joint actuator torques and the resulting motion. The dynamic equations of motion form a basis of many computations in mechanical design, controls and simulation \cite{c9}. This equation of motion consists of joint space positions, velocity, acceleration and force vectors.The common form of a serial manipulator's dynamical model is in the following state space form.

\begin{equation}
M(q)\ddot{q} + C(q,\dot{q}) + G(q) = \tau \\
\end{equation}

In the equation, $q$ denotes the vector of joint angles, $M(q)$ is the inertia matrix, $C(q,\dot{q})$ denotes the Coriolis matrix and the $G(q)$ represents the gravitational vector. The equation equates to the vector of actuator torques. There are two approaches towards the dynamical modeling of a system, the Euler-Lagrange approach and the Newton-Euler approach. In Euler-Lagrange model, the links are considered together and the model is obtained analytically using kinetic and potential energy of the system.The Newton-Euler method forms an equation with recursive solution. The Forward Kinematics equation defines a function between space of Cartesian poses and the space of joint poses. The velocity relationship are determined by the Jacobian of this function \cite{c10}. 

\section{Background}

The first step to controlling a robot is to understand the mathematical model of the system. Baxter has been designed to operate safely in collaboration with humans, and, as a result, has become one of the most used research platforms in robotics labs. Consequentially, many groups has performed research on Baxter's kinematics and dynamics. Silva, et al., explored the forward kinematics of Baxter's arms, and the workspace defined by the arm span \cite{c1}. Smith, et al., derived dynamic equations for the arms, validated by measuring torques throughout pre-defined trajectories \cite{c2}. Silva and Ju, et al. also derived kinematic models of the arms and modeled them in MATLAB \cite{c3}.

The forward kinematics of a serial manipulator is a very well-established concept in robotics research, the most commonly used technique being Denavit-Hartenberg parameters \cite{c4}. Inverse Kinematics is more complicated, as there is no universal mechanism to derive the inverse kinematics equations of a manipulator. In fact, closed-form solutions do not even exist for redundant (7-DOF or more) manipulators. Most work on inverse kinematics of redundant robots focus on iterative or numerical approaches \cite{c5} \cite{c6} \cite{c7}, which suffer from their own share of issues. Dr. Williams II of Ohio University penned an excellent article discussing the 7-DOF Baxter arm kinematics \cite{c8}.

The dynamics of Baxter's arm has been a subject of research too, albeit less than its kinematics. The mathematical model of Baxter's dynamics is extremely complicated, with over ``half a million coefficients'', as reported by Yang, et al. \cite{c9}. ``Advanced Technologies in Modern Robotics Applications'' \cite{c10} also discusses the Euler-Lagrange method for dynamical modeling of Baxter, which we will use as our primary reference for this purpose.

\section{Kinematics}

The Baxter robot is endowed with a proficient SDK provided by its makers, Rethink Robotics. The native support uses Gazebo as the simulation environment. Moreover, it has several interfaces for tele-operation of the simulation model as well as the real-life model of Baxter. Here, we discuss our approaches to solve the various expected and reach goals that we have put forth for the project. The various subsections discuss the methodology, tools and resources that we used.

\subsection{Denavit-Hartenberg Parameters}

DH parameters are a set of four variables that define the spatial relationship between two valid coordinate frames. These variables are $d$ (translation along old z), $\theta$ (rotation about old z), $a$ (translation along new x) and $\alpha$ (rotation about new x). These parameters are combined using the DH framework to derive the transformation matrix that links the two coordinate frames through matrix multiplication.

\begin{center}
$T^{n-1}_{n} = Rot_{z}(\theta) \cdot Trans_{z}(d) \cdot Trans_{x}(a) \cdot Rot_{x}(\alpha)$
\end{center}

DH parameters are entirely dependent on the way the frames are assigned on the manipulator. The figure below depicts our frame assignment for the Baxter left arm, keeping in line with the DH convention. The table below shows the resultant DH parameters as calculated for the Baxter robot's left arm. The numerical values of the link lengths were taken from official Baxter documentation by Rethink Robotics.

\begin{figure}[!h]
\centering
\includegraphics[width=\columnwidth]{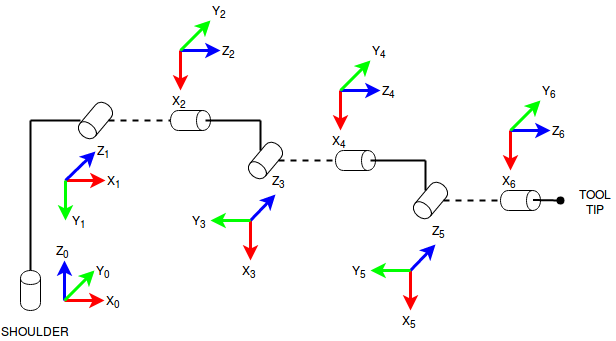}
\caption{Link Diagram with coordinate frame assignment}
\label{fig:baxter_frame} 
\end{figure}

\begin{table}[h]
\caption{DH PARAMETERS FOR BAXTER LEFT ARM}
\label{table b}
\begin{center}
\begin{tabular}{|c|c|c|c|c|}
\hline
\thead{\textbf{Link}} & \thead{\bf{d} (m)} & \thead{\bf{$\theta$}  (rad) }  & \thead{\bf{a}  (m)} & \thead{\bf{$\alpha$} (rad)} \\
\hline
S0-S1  & 0.27035  & $\theta_{0}$ & 0.069 & $-\pi/2$    \\
\hline
S1-E0  & 0 & $\theta_{1} + \pi/2$ & 0 & $\pi/2$\\
\hline
E0-E1  & 0.36435 & $\theta_{2}$ & 0.069 & $-\pi/2$\\
\hline
E1-W0  & 0 & $\theta_{3}$ & 0 & $\pi/2$\\
\hline
W0-W1  & 0.37429 & $\theta_{4}$ & 0.01 & $-\pi/2$\\
\hline
W1-W2  & 0 & $\theta_{5}$ & 0 & $\pi/2$\\
\hline
W2-EE  & 0.229525 & $\theta_{6}$ & 0 & 0\\
\hline
\end{tabular}
\end{center}
\end{table}

\subsection{Forward Pose Kinematics}

Baxter's forward pose kinematics (FPK) equations give the 6-DOF pose (3 position and 3 orientation) of the end-effector as a function of the seven joint angles of the arm manipulator. Computing the DH parameters is the first step towards deriving the closed-form FPK equations. 

Forward kinematics is calculated using transformation matrices. For the arm in question, the transformation matrix from the base to the tip frame using the several intermediate transformation matrices is given as:

\begin{center}
$T^0_{7} = T^0_{1} \cdot T^1_{2} \cdot T^2_{3} \cdot T^3_{4} \cdot T^4_{5} \cdot T^5_{6} \cdot T^6_{7}$  
\end{center}

The transformation matrices used in the above equation make use of the DH parameters as given in the generalized formula for transformation matrix $T^{n-1}_{n}$. The final end-effector position equations as a function of the joint angles, i.e. the FPK equations are given below.

\begin{center}
$X_{ee}^{b} = d_1.c_0 - l_4(c_5(s_0s_2s_3 - c_0c_1c_3 + c_0c_2s_1s_3) + s_5(c_4(c_0c_1s_3 + c_3s_0s_2 + c_0c_2c_3s_1) + s_4(c_2s_0 - c_0s_1s_2))) - l_3(s_0s_2s_3 - c_0c_1c_3 + c_0c_2s_1s_3) - d_3.s_4(c_2s_0 - c_0s_1s_2) + l_2.c_0c_1 - d_2.s_0s_2 - d_3.c_4(c_0c_1s_3 + c_3s_0s_2 + c_0c_2c_3s_1) + d_4.c_6(s_5(s_0s_2s_3 - c_0c_1c_3 + c_0c_2s_1s_3) - c_5(c_4(c_0c_1s_3 + c_3s_0s_2 + c_0c_2c_3s_1) + s_4(c_2s_0 - c_0s_1s_2))) + d_4.s_6(s_4(c_0c_1s_3 + c_3s_0s_2 + c_0c_2c_3s_1) - c_4(c_2s_0 - c_0s_1s_2)) - d_2.c_0c_2s_1$
\end{center}

\begin{center}
$Y_{ee}^{b} = l_4.(c_5(c_1c_3s_0 + c_0s_2s_3 - c_2s_0s_1s_3) - s_5(c_4(c_1s_0s_3 - c_0c_3s_2 + c_2c_3s_0s_1) - s_4(c_0c_2 + s_0s_1s_2))) + d_1.s_0 + l_3(c_1c_3s_0 + c_0s_2s_3 - c_2s_0s_1s_3) + d_4.s_6(s_4(c_1s_0s_3 - c_0c_3s_2 + c_2c_3s_0s_1) + c_4(c_0c_2 + s_0s_1s_2)) + d_3s_4(c_0c_2 + s_0s_1s_2) + d_2.c_0s_2 + l_2.c_1s_0 - d_3.c_4(c_1s_0s_3 - c_0c_3s_2 + c_2c_3s_0s_1) - d_4.c_6(s_5(c_1c_3s_0 + c_0s_2s_3 - c_2s_0s_1s_3) + c_5(c_4(c_1s_0s_3 - c_0c_3s_2 + c_2c_3s_0s_1) - s_4(c_0c_2 + s_0s_1s_2))) - d_2.c_2s_0s_1$
\end{center}

\begin{center}
$Z_{ee}^{b} = l_1 - l_2.s_1 - d_2.c_1c_2 - l_3.c_3s_1 - l_3.c_1c_2s_3 - l_4.c_3c_5s_1 + d_3.c_1s_2s_4 + d_3.c_4s_1s_3 - d_3.c_1c_2c_3c_4 - l_4.c_1c_2c_5s_3 + d_4.c_1c_4s_2s_6 + d_4.c_3c_6s_1s_5 + l_4.c_1s_2s_4s_5 + l_4.c_4s_1s_3s_5 - d_4.s_1s_3s_4s_6 - l_4.c_1c_2c_3c_4s_5 + d_4.c_1c_2c_3s_4s_6 + d_4.c_1c_2c_6s_3s_5 + d_4.c_1c_5c_6s_2s_4 + d_4.c_4c_5c_6s_1s_3 - d_4.c_1c_2c_3c_4c_5c_6$
\end{center}

The above equations can be used directly to get the end effector positions by feeding in the inputs for the joint angles. The terms $c_i$ and $s_i$ in the equations denote $cos(\theta_i)$ and $sin(\theta_i)$ respectively, while $l_i$ and $d_i$ refer to the lengths and offsets at each set of links. Note that $d_4$ for Baxter is zero.
\\
\\
\textbf{Results} \\

The performance obtained by our Forward Kinematics setup has comparable results with that of the PyKDL library developed for the Baxter arm by Rethink Robotics itself. Comparing performances over 5 random poses for the PyKDL standard library and our MyKDL package, the resultant performance results obtained were:

\begin{flushleft}
Average Position Error = $0.00008$ m \\
Average Orientation Error = $0.00017$ deg
\end{flushleft}

\subsection{Skeleton Model}

\begin{figure}[!h]
\centering
\includegraphics[width=\columnwidth]{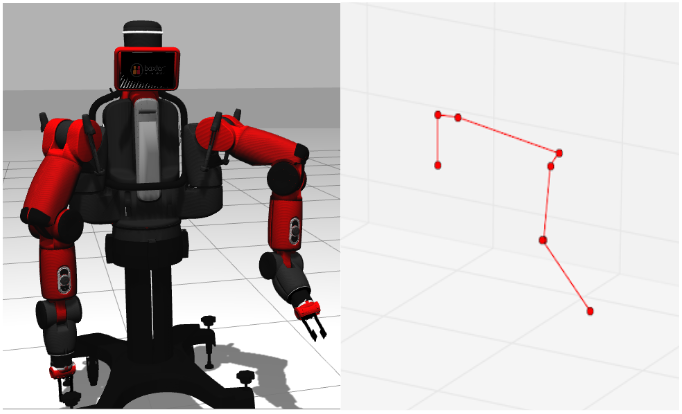}
\caption{Baxter Arm represented by a skeleton model}
\label{fig:skeleton}
\end{figure}

In order to visualize our conceived methodology and test its working, we developed a skeletal model of the Baxter arm. Based on the DH parameters, we obtained a 3D stick diagram of the arm using the \textit{matplotlib} library in Python, showing the Cartesian configuration (position and orientation) of the various links in a given joint space configuration. We ensured its credibility by visually comparing the same with the Baxter model in simulation provided by Rethink Robotics. Figure \ref{fig:skeleton} show the Baxter arm in the home configuration in Gazebo and corresponding visualization of our skeletal model.

\subsection{ Workspace Analysis} 

Baxter arm's redundancy and kinematic structure enables it to get rid of singularities, evidently making it capable of having a large workspace. Despite the joints of Baxter not having full 360 degree range of motion, the seven degrees of freedom and the presence of a non-spherical wrist make it possible for the robot to eventually get to any point within the bounding canopy formed by maximum reach position of the EE over complete range of motion for the joints. 

\begin{figure}[!h]
\centering
\includegraphics[width=\columnwidth]{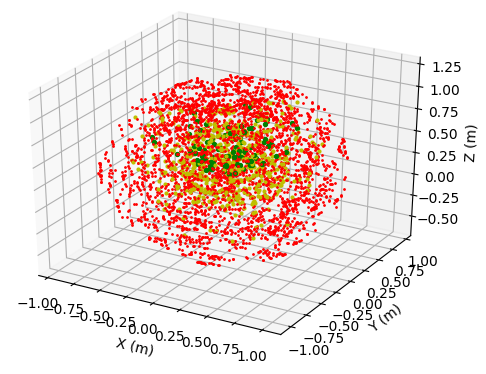}
\caption{Workspace Diagram of Baxter Left Arm, colored based on Yoshikawa Manipulability Index (Red = Low,  Yellow = Medium,  Green = High)}
\label{fig:workspace_diag} 
\end{figure}

We exploit the developed skeletal model to obtain the canopy of end-effector positions that encase the complete workspace. However, since we weren't able to incorporate the self-collision space in the rudimentary skeletal model, the resultant workspace tends to have those regions that are practically not reachable by the Baxter arm. Figure \ref{fig:workspace_diag} shows the workspace derived by our model.

\subsection{Jacobian Matrix}

While forward position kinematics enables us to calculate the pose of the end-effector at all times, velocity kinematics is much more useful and powerful in the context of manipulator control. Velocity kinematics relates the joint velocities and the end-effector velocities. Interestingly, although the forward pose kinematics of a robot manipulator is non-linear, the velocity kinematics is always linear. 

The established approach to mapping the velocities in the joint space and task space is the Jacobian matrix. The Jacobian is a $n\times m$ matrix (where $n$ is the dimension of the task space and $m$ is the joint space dimension), computed through the partial differentiation of the FPK equations, as shown below.

\begin{equation}
J(q) = \begin{bmatrix}
        A(q) \\
        B(q) \\
        \end{bmatrix} = 
        \begin{bmatrix}
        \frac{\partial x_t(q)}{\partial q_1} & 
        \frac{\partial x_t(q)}{\partial q_2} & ... & 
        \frac{\partial x_t(q)}{\partial q_m} \\
        \xi_1 z_1 & \xi_2 z_2 & ... & \xi_m z_m
        \end{bmatrix}
\end{equation}

Here, $x_t(q)$ is the $n\times1$ vector-valued FPK equation, $z_k$ is the $k^{th}$ joint rotation axis in the base frame and $\xi_k$ represents the type of the $k^{th}$ joint ($0$ for prismatic and $1$ for revolute). 

The arm manipulator of the Baxter research robot has 7 joints, all of them revolute. Hence, the joint space dimension $m$ is 7 and the task space dimension $n$ is 6 while all $\xi$s are 1. For the purpose of this project, computing the Jacobian matrix for Baxter has been done through MATLAB Symbolic Math Toolbox since it is extremely complicated to compute by hand.

\subsection{Velocity Kinematics}

The Jacobian matrix is central to the velocity kinematics of a serial manipulator. Let $x$ represent the $6\times1$ end-effector pose in task space and $q$ refer to the $7\times1$ vector of joint angles. Then, $\dot{x}$ and $\dot{q}$ are the velocities in the task space and the joint space respectively. The Jacobian matrix $J(q)$ relates the two velocities as follows: 

\begin{equation}
\dot{x} = J \dot{q}\\
\end{equation}
\begin{equation}
\dot{q} = J^{\dagger} \dot{x}
\end{equation}

Here, $J^{\dagger} = J^{T}(JJ^{T})^{-1}$ is the Moore-Penrose pseudoinverse of $J$, a more generalized matrix inverse for non-square matrices. The above equations form the set of velocity kinematics equations, the first representing the forward velocity kinematics (FVK) while the second is the inverse velocity kinematics (IVK). 

The Jacobian matrix $J(q)$ plays a big role in the kinematic and dynamic analysis of a serial manipulator, hence, computing it is an important step. Since it relates the task space velocities and the joint space velocities, the Jacobian can be used to measure the \textit{manipulability} of a joint pose, i.e. the ability to apply forces or velocities in different directions. 

Singularities are joint poses which do not allow for task space velocities in certain directions, essentially reducing the effective degrees of freedom of the manipulator. The Jacobian provides a simple way to measure the closeness of a joint pose to a singularity, thereby, providing ample warning to stay away from such configurations. Such measures are called \textit{manipulability indices}. 

One of the most widely-used is the \textit{Yoshikawa's manipulability index}, defined as $\sqrt{|JJ^T|}$ which goes to zero as the manipulability of the joint pose decreases. As part of our library implementation, we compute the Yoshikawa's index whenever we (re)compute the Jacobian matrix. This enables us to warn the users whenever Baxter's joint pose is approaching a singular configuration.

Jacobian also plays a big role in redundancy resolution in inverse kinematics of redundant manipulators. The null projector matrix of the Jacobian maps an arbitrary vector to the nullspace of the Jacobian, which means, we can use the null projector matrix in the inverse velocity kinematics equation to move the joints into a more desired position without effecting the task space velocity. The null projector matrix is defined as $I - J^{\dagger}J$ and this redundancy resolution technique is summarized in the equation below.

\begin{equation}
\dot{q} = J^{\dagger} \dot{x} + (I - J^{\dagger}J)\dot{q}_{r}
\end{equation}

Our library provides support for all these important concepts. The implementation updates the Jacobian and can compute its pseudo-inverse, manipulability index and null projector matrix so that users can leverage these properties to build more robust controllers.

\subsection{6-DOF Inverse Pose Kinematics}

Before going towards control of the 7-DOF redundant Baxter arm, we first work on only 6 degrees of freedom by freezing the joint E0. In this section, we perform inverse kinematics on 6-DOF Baxter robot arm. As we have locked the joint E0, we have $\theta_3 = 0$. Additionally, we take another reasonable approximation of $L_5 = 0$. We also make use of a new parameter $L_h$ to simplify the results, defined as

\begin{equation}
L_h = \sqrt{L_2^2 + L_3^2}
\end{equation}

We perform the forward position kinematic of the 6-DOF arm. This is got by substituting the values as above in the FPK equations. We get the end-effector pose $T_6^0$ from the matrix as follows.

\begin{equation}
           T_6^0 =\begin{bmatrix}
                   r_{11} & r_{12} & r_{13} & x_6^0 \\
                   r_{21} & r_{22} & r_{23} & y_6^0\\
                   r_{31} & r_{32} & r_{33} & z_6^0 \\
                   0 & 0 & 0 & 1
                   \end{bmatrix}
\end{equation}      

Our 6-DOF analytical solution basically maps from this Transformation matrix $T_6^0$ to the joint angles {$\theta_1, \theta_2, \theta_4, \theta_5, \theta_6, \theta_7$}. The position vector in this case is given as 

\begin{equation}
\{P_6^0\} = \begin{bmatrix} x_6^0 \\ y_6^0 \\ z_6^0 \end{bmatrix} = \begin{bmatrix}
                  c_1(L_1 + L_h c_2 + L_4 c_{24}) \\
                  s_1(L_1 + L_h c_2 + L_4 c_{24} ) \\
                         - L_h s_2 - L_4 s_{24}
                     \end{bmatrix}  \\ 
\end{equation}

Therefore the joint angles $\theta_1, \theta_2, \theta_4$ are solved analytically as follows

\begin{center}
$ \theta_1 = atan2(y_4^0 , x_4^0) $\\~\\ 
$ \theta_{2_{1,2}} = 2 \tan^{-1}(t_{1,2}) $\\~\\
$\theta_{4_{1,2}} = atan2(-z_6^0 -L_h s_{2_{1,2}} , \frac{x}{c_1} - L_1 - L_h c_{2_{1,2}}) - \theta_{2_{1,2}}$ \\~\\
\end{center}
            
where the variables $E, F, G$  are

\begin{center}
$ E = 2L_h (L_1- \frac{x}{c_1})$  \\~\\
            $F = 2 L_h z$ \\~\\
            $G = (\frac{x^2 }{c_1^2} + L_h^2 + L_1^2 - L_4^2 + z^2 - 2 \frac{L_1 x}{ c_1}) 
			$ \\~\\
\end{center}

and the variable $ t_{1,2} $ is

\begin{center}
			$ t_{1,2} = \frac{-F \pm \sqrt{E^2 + F^2 - G^2}}{G-E}$ \\ 
\end{center}

Therefore, we get two possible solutions for the angles from the translational components as shown in the table below.

\begin{center}
\begin{tabular}{|c|c|c|c|} 
\hline
Sol. 1 & $\theta_1$ & $\theta_{2_{1}}$ & $\theta_{4_{1}}$ \\
\hline
Sol. 2 & $\theta_1$ & $\theta_{2_{2}}$ & $\theta_{4_{2}}$ \\
\hline
\end{tabular}
\end{center} ~\\

These two solutions share the same common $\theta_1$ and $(\theta_2,\theta_4)$ pairs corresponding to the elbow up and elbow down configurations.

Now for the solutions of the joint angles $\theta_5, \theta_6$ and $\theta_7$ we take use of the rotational matrix $R_6^0$ and the previous results $(\theta_1, \theta_2, \theta_4)$ into consideration. Since $\theta_5, \theta_6$ and $\theta_7$ are found only in the $R_6^3$, we find $R_6^3$ from $R_6^0$ as follows:

\begin{center}
$R_6^3(\theta_5,\theta_6,\theta_7) = [R_3^0(\theta_1,\theta_2,\theta_4)]^T [R_6^0] = \begin{bmatrix}
                   R_{11} & R_{12} & R_{13} \\
                   R_{21} & R_{22} & R_{23}\\
                   R_{31} & R_{32} & R_{33}\\
                   \end{bmatrix} $
\end{center}

\begin{center}
$R_6^3(\theta_5,\theta_6,\theta_7) =\begin{bmatrix}
                   -s_5 s_7 + c_5 c_6 c_7 & -s_5 c_7 -c_5 c_6 s_7 & c_5 s_6 \\
                   s_6 c_7 & -s_6 s_7 & -c_6\\
                   c_5 s_7 +s_5 c_6 c_7 & c_5 c_7 - s_5 c_6 s_7 & s_5 s_6\\
                   \end{bmatrix} $
\end{center}

Therefore, using $R_6^3$, we can find $\theta_5, \theta_6$ and $\theta_7$ from the equations given below.

\begin{center}
$\theta_5 = atan2(R_{33},R_{13}) $ \\~\\
$\theta_6 = atan2(\frac{R_{21}}{c_7},-R_{23}) $ \\~\\
$\theta_7 = atan2(-R_{22},R_{21})$ \\~\\
\end{center}

\subsection{Iterative Solvers for Redundant IPK}

Although we have a valid IK solution derived for the 6-DOF arm by locking a joint, this does not leverage the redundancy in the arm to find better solutions than the one returned by 6-DOF IK. On the other hand, 7-DOF IK solutions can not be derived analytically and numerical iterative solvers are the common approach to this problem. We implement a few standard iterative IK solvers as part of our library and describe them below.
\\
\subsubsection{Jacobian Pseudoinverse}

Jacobian PseudoInverse technique is the most basic of iterative IK solvers and is often treated as the baseline method. Computation of the pseudoinverse for non-square Jacobian matrices gives us the inverse velocity kinematics for the robot. The position kinematics is then obtained by integrating the velocity kinematics over several time steps. The pseudoinverse approach to iterative IK starts with taking the joint angle positions for the current configuration as the seed angles for integration over time. 

The algorithm is run repeatedly with a value of $\dot{x}$ taken as a small vector in the direction of the vector joining the current end-effector position to the target end-effector position. Thereafter, using the obtained Jacobian Pseudoinverse, we compute the joint angular velocity. Integrating it over a constant value of time-step and comparing the obtained final Cartesian pose with the target pose, we gradually converge to the latter. This process of computation and comparison until convergence makes this approach an iterative technique. Algorithm 1 shows the pseudo-code for this algorithm.
 
\begin{algorithm}
\caption{Pseudoinverse Method}
\textbf{PROCEDURE} PseudoInverse($x_{des}, q_{seed}, step$) \\
$x \leftarrow FPK(q_{seed})$ \\
$q \leftarrow q_{seed}$ \\
\begin{algorithmic}[0]
\REPEAT
\STATE
$\Delta x \leftarrow x_{des} - x$ \\
$\dot{x} \leftarrow \frac{\Delta x}{||\Delta x||} \times step$ \\
$q \leftarrow q + J^{\dagger}(q)\dot{x}$ \\
$x \leftarrow FPK(q)$ \\
\UNTIL{$||x_{des}-x|| < \epsilon$}
\end{algorithmic}
\end{algorithm}

\subsubsection{Pseudo Inverse with Random Restarts}

One issue with the Jacobian Pseudo Inverse is that it does not work well when the arm manipulator has joint limits, which Baxter's arm does. Hence, we observed a very low solve rate with the vanilla Pseudo Inverse technique. The solution to this is to use random restarts. The algorithm works in essentially the same way as Jacobian PseudoInverse except that, whenever the solver hits a joint limit, it randomly restarts the joint pose and attempts again. This has led to a drastic improvement in the solve rate of the IK Solver for the Baxter arm. Hence, Pseudo Inverse with Random Restarts is our library's default iterative IK Solver.

Video showing the implementation of the Pseudo Inverse IK technique is available here: \url{https://www.youtube.com/watch?v=E54lb_UORLA}

\begin{figure}[!h]
\centering
\includegraphics[width=0.75\columnwidth]{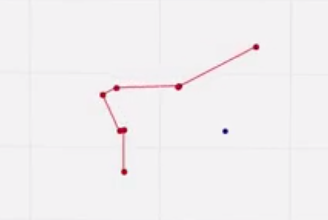}
\caption{PseudoInverse-RR in action}
\label{fig:pinv_solve} 
\end{figure}

\subsubsection{Cyclic Coordinate Descent}

Inverse kinematics for serial manipulators has always been a challenging task. It has been an area of research for long, especially for redundant robots with more than 6 degrees of freedom with no closed-form solution. The proposed task is more challenging when the number of variables (joint angles) is less than the number of equations derived from the input pose and orientation of the robot.

The Cyclic Coordinate Descent method iteratively tries to make the end-effector first converge onto a sphere with radius equaling the distance between the base of the manipulator and the end-effector and thereafter making the same converge onto the target position. 

The variation made in joint angle values for joints farthest from the base does not reflect on the complete chain. Such joints are tried first while gradually moving towards joints proximal to the base. Since it does not involve the Jacobian matrix, this method is free from issues of matrix inversion and is also free from singularities. Figure \ref{fig:ccd} shows the flowchart for CCD.

The proposed technique indeed comes with some caveats like inability to converge to a defined orientation and inherent computational limitations while each iteration tries to make the end-effector fall on the line joining the joint be actuated and the target position.

\begin{figure}[!h]
\centering
\includegraphics[width=0.75\columnwidth]{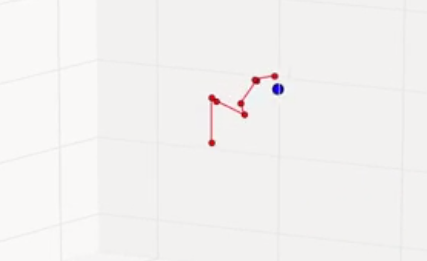}
\caption{CCD IK Solver in action}
\label{fig:ccd_solve} 
\end{figure}

Video showing the implementation of the same is available here: \url{https://www.youtube.com/watch?v=Wy0hyKiDvaw}

\begin{figure}[!h]
\centering
\includegraphics[width=8cm]{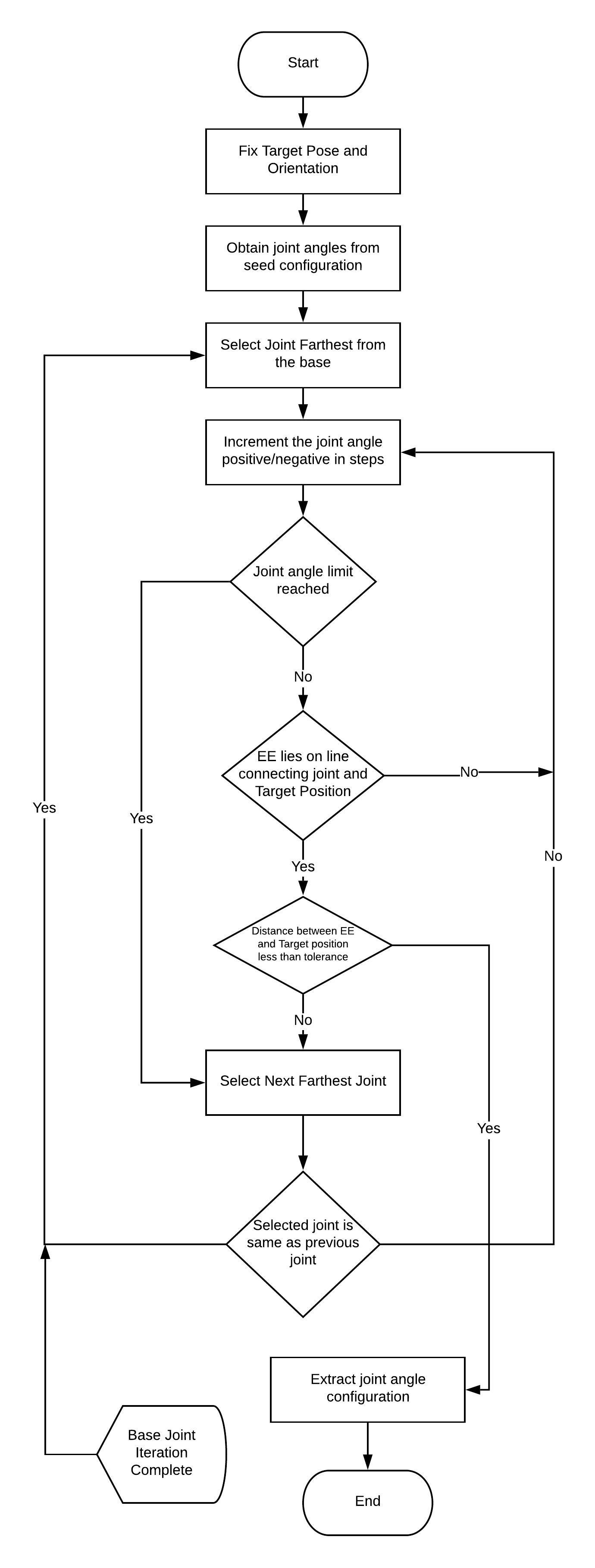}
\caption{CCD Algorithm Implementation}
\label{fig:ccd} 
\end{figure}    

\section{Dynamics}

The mass-inertia matrices, Coriolis effect coefficients and gravity coefficients are three major components needed to define the dynamical aspects a robot. Low level controllers like a PID controller used for trajectory tracking and set-point tracking work without incorporating the dynamical model of the manipulators. 

However, force/torque control on the robot as well as handling external interactions with the manipulator need a dynamical model of the same to ensure that the position/trajectory/constraints are tracked regardless of any variation in inherent parameters over the course of motion. Manipulator control techniques like computed torque control,  adaptive control, robust control and impedance all incorporate a parameterized dynamical model of the robot that controls the output positions and forces/torques in joint as well as task space. 

Thus, the M, C and G matrices are determined to ascertain the dynamical model of the Baxter arm in real-time and implement appropriate control strategies. 

\subsection{The Euler-Lagrange Approach}

The Euler-Lagrangian formulation to derive the dynamical model of a robotic manipulator incorporates the computation of Kinetic and Potential energies of the system parameterized by the joint angles, i.e. the orientation of the arm. Thereafter, computation of the Lagrangian and its derivatives with time and joint variables results in the required dynamic model. Equations 4, 5 and 6 elaborate over this methodology. 

The equations for Kinetic and potential energy of the robotic manipulator based on the Uicker/Kahn formulation that considers rotational matrices between joints instead of joint angular velocities to compute the former are given as:
\begin{equation}
K = \frac{1}{2} \sum_{i=1}^{n} \sum_{j=1}^{i} \sum_{k=1}^{i} [Tr(U_{ij} J_i U_{ik}^T)  \dot{q_j} \dot{q_k}]
\end{equation}
\begin{equation}
P = \sum_{i=1}^{n} -m_i g (T_{0}^{i} \bar{r_i} )
\end{equation}

where the matrix $U_{ij}$ representing the rate of change of points on link $i$ relative to the base as the joint position $q_j$ changes is given as:

\[
    U_{ij} \equiv \frac{\partial T_i^0}{\partial q_i} =
\begin{cases}
    T _{j-1}^{0} Q_j T_{j-1}^i &  j\leq 1 \\
    0            & j > 1
\end{cases}
\]

Further, the Lagrangian and subsequent joint-torque calculation is given as:
\begin{equation}
L = K - P
\end{equation}
\begin{equation}
\tau = \frac{d}{dt} (\frac{\partial L}{\partial \dot{q}}) - \frac{\partial L}{ \partial q} 
\end{equation}

\subsection{Mass Matrix - M}  
The inertia tensor matrix for an individual link in the local frame is given as:

\begin{equation}
I = 
\begin{bmatrix}
I_{xx} & I_{xy} & I_{xz} \\
I_{yx} & I_{yy} & I_{yz} \\
I_{zx} & I_{zy} & I_{zz} \\
\end{bmatrix}
\end{equation}

The Baxter robot URDF provided by Rethink Robotics already provides these inertia tensor values for each individual links. The proposed library exploits these values and sequentially converts them to the base frame to compute the Kinetic energy needed for the computation of Lagrangian.
On processing equations (10) and (11) and transformation of the inertia tensors to the base frame, the final equation for individual elements of the Mass matrix is given as:
\begin{equation}
M_{i,k} = \sum_{j=max(i,k)}^{n} Tr(U_{jk} J_j U_{ji}^T) \hspace{10mm} i,k = 1,2 ...n
\end{equation}

where, 

\begin{equation}
J_{i} = 
\begin{bmatrix}
\frac{-I_{xxi} + I_{yyi} + I_{xxi}}{2} & I_{xyi} & I_{xzi} & m_{i}\bar{x_i} \\
I_{xyi} & \frac{-I_{xxi} + I_{yyi} + I_{xxi}}{2} & I_{yzi} & m_{i}\bar{y_i} \\
I_{xzi} & I_{yzi} & \frac{-I_{xxi} + I_{yyi} + I_{xxi}}{2} & m_{i}\bar{z_i} \\
m_{i}\bar{x_i} &m_{i}\bar{y_i} & m_{i}\bar{z_i}  & m_{i} \\
\end{bmatrix}
\end{equation}
 
\subsection{Coriolis Matrix - C}

Coriolis effect is generally neglected for smaller systems to merely circumvent the intricacies involved with the solution. However, for a serial manipulator with several links like for Baxter, we can't neglect the Coriolis effect. 

The Coriolis Matrix equation is given as:

\begin{equation}
C_i = \sum_{k=1}^{n} \sum_{m=1}^{n} h_{ikm} \dot{q_k} \dot{q_m}
\end{equation}

\begin{equation}
h_{ikm} = \sum_{j=max(i,k,m)}^{n} Tr( U_{jkm} J_j U_{ji}^T )
\end{equation}

where the derivation of the interaction between the joints $U_{jkm}$ is given as:

\[
    U_{ijk} \equiv \frac{\partial U_{ij}}{\partial q_k} =
\begin{cases}
    T_{j-1}^{0} Q_j T _{k-1}^{j-1} Q_k T _{i}^{k-1} &  i\geq k\geq j \\
    T_{k-1}^{0} Q_k T _{j-1}^{k-1} Q_j T _{i}^{j-1} &  i\geq j \geq k\\
    0 &  i $<$ j or j $<$ k \\
\end{cases}
\]

where for Baxter, as all the joints are revolute, Q is given as: 

\begin{equation}
Q_ = 
\begin{bmatrix}
0 & -1 & 0 & 0 \\
1 & 0 & 0 & 0 \\
0 & 0 & 0 & 0 \\
0 & 0 & 0 & 0 \\
\end{bmatrix}
\end{equation}

\subsection{Gravity Matrix - G}

For a 3D manipulator arm, each mass point is acted upon by the force of gravity that translates to certain resultant gravitational force acting on all the individual joints. The G matrix is a 3 X 1 vector that represents the dynamically varying gravity force that acts on these joints. 

The G matrix is finally given as:
\begin{equation}
G_i = \sum_{j=1)}^{n} -m_j g U_{ji} \bar{r_j}
\end{equation}

where \textbf{g} = [0 0 -9.81 0] is the gravity row vector.

\subsection{Trajectory Following}

The Baxter robot is supported in several simulation environments like Gazebo, V-REP and Klamp't with almost all features supported. However, the official SDK provided by makers Rethink Robotics supports Gazebo simulation environment with communication over ROS as publisher/subscriber technique or as services. Moreover, the SDK also supports several communication interfaces with external control as well as kinematics libraries to be used as plugins. A similar feature provided in the SDK is the \textit{baxter\_interface } repository that holds the Python API for interacting with the Baxter Research Robot, in simulation as well as with the real robot. This comprises of a set of classes that provide wrappers around the ROS communications from Baxter, allowing for direct python control of the different interfaces of the robot.

To test our results for the Forward and Inverse Kinematics, we generated a set of way-points over a predefined path at very short-intervals and used our IK Pseudo-inverse IK service to generate corresponding joint-angles for the motion. To simulate the same motion on the robot, we used the Baxter Interface that defines each arm and individual gripper of the robot as separate entities using a \textit{limb} class. We feed our set of joint angles for each way-point to the same and using the in-built controller, the Baxter SDK simulates the robot arm along our trajectory. 

Video showing the implementation of the trajectory tracking setup is available here: \url{https://www.youtube.com/watch?v=PPKZ8XThTAk}

\begin{figure}[!h]
\centering
\includegraphics[width=\columnwidth]{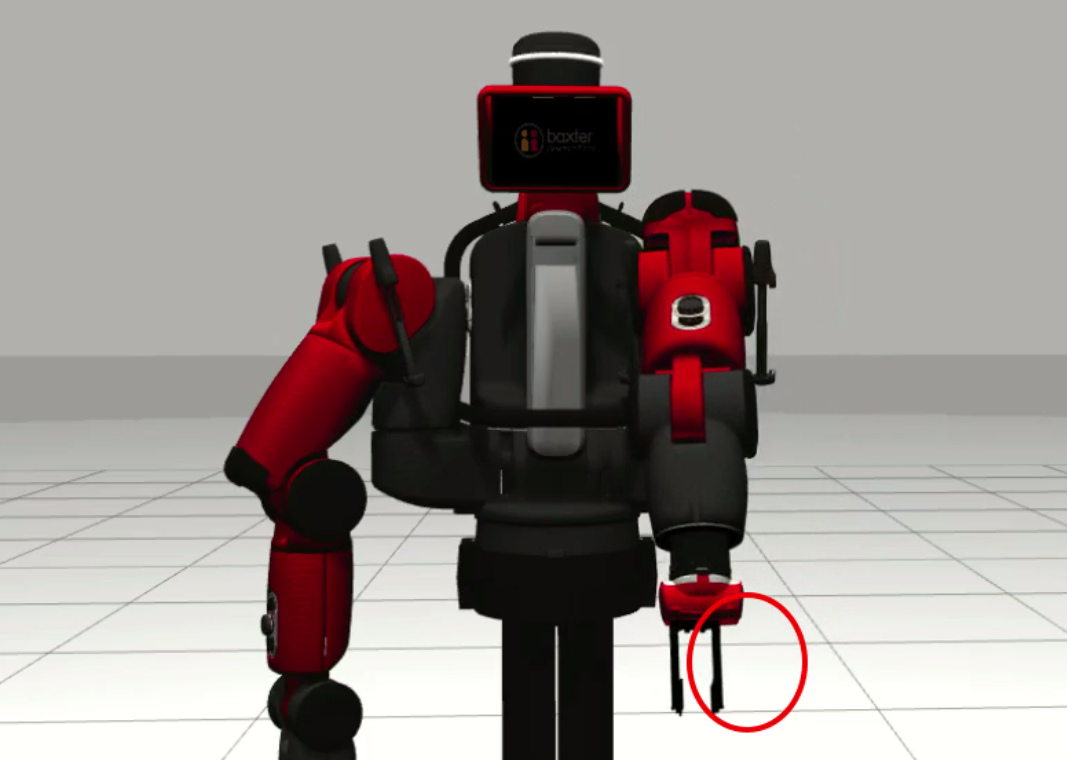}
\caption{Baxter Simulation following a circular trajectory. It resolves the way points into joint space with our IK Solver}
\label{fig:traj_follow} 
\end{figure}

Experimental results (Fig. \ref{fig:traj_follow}) show comparable real-time performance of the proposed IK service and strengthen our argument that given nearby Cartesian positions as start and end poses for one IK solution, our proposed library has good performance.

\section{Conclusion}

We developed own version of a Kinematics and Dynamics library for Baxter Robots arm which provided more functionalities over the originally available SDK and the PyKDL library from Rethink Robotics. Our library contains implementations of Forward Kinematics and iterative algorithms for 7-DOF Inverse Kinematics with 6-DOF solution as a backup if 7-DOF IK fails. It provides M, C, and G matrices for the dynamics of the Baxter arm which can further be used in development of better and efficient control algorithms. We also support redundancy resolution of IK solutions by providing the nullspace projector matrix of the Jacobian, along with many other functions which play a role in the kinematic and dynamical analysis of Baxter. Our hope is that this toolbox is a step towards making Baxter a mush more user-friendly robot to research on and work with.

\section{Future Work}

While implementing the above mentioned techniques for inverse kinematics, many fascinating ideas in iterative IPK struck our mind which we plan to implement in future. Some of them are null space optimization, combination of CCD and Pseudo-Inverse, and Joint Space RRT with Pseudo-Inverse. We also plan to implement some motion planning component in the library with better ROS support for the robotics community to use.

%%%%%%%%%%%%%%%%%%%%%%%%%%%%%%%%%%%%%%%%%%%%%%%%%%%%%%%%%%%%%%%%%%%%%%%%%%%%%%%%


\begin{thebibliography}{99}

\bibitem{c1} L. E Silva, T. M. Tennakoon, M. Marques, and A. M. Djuric, “Baxter Kinematic Modeling, Validation and Reconfigurable Representation,” 2016, vol. 2016–April, no. April.
\bibitem{c2} A. Smith, C. Yang, C. Li, H. Ma, and L. Zhao, “Development of a dynamics model for the Baxter robot,” in 2016 IEEE International Conference on Mechatronics and Automation, IEEE ICMA 2016, 2016, pp. 1244–1249.
\bibitem{c3} Z. Ju, C. Yang, and H. Ma, “Kinematics modeling and experimental verification of baxter robot,” in Proceedings of the 33rd Chinese Control Conference, 2014, pp. 8518–8523.
\bibitem{c4} R. S. Hartenberg J. Denavit "A kinematic notation for lower pair mechanisms based on matrices" J. Appl. Mech. vol. 77 no. 2 pp. 215-221 Jun. 1955. 
\bibitem{c5} Kazerounian, K. (1987). On the numerical inverse kinematics of robotic manipulators. Journal of mechanisms, transmissions, and automation in design, 109(1), 8-13.
\bibitem{c6} Beeson, P., \& Ames, B. (2015, November). TRAC-IK: An open-source library for improved solving of generic inverse kinematics. In Humanoid Robots (Humanoids), 2015 IEEE-RAS 15th International Conference on (pp. 928-935). IEEE.
\bibitem{c7} Aristidou, A., \& Lasenby, J. (2011). FABRIK: a fast, iterative solver for the inverse kinematics problem. Graphical Models, 73(5), 243-260.
\bibitem{c8} R.L. Williams II, “Baxter Humanoid Robot Kinematics”, Internet Publication,     
https://www.ohio.edu/mechanical-faculty/williams/html/pdf/BaxterKinematics.pdf, April 2017. 
\bibitem{c9} Yang, C., Ma, H., \& Fu, M. (2016). Advanced technologies in modern robotic applications. Springer Singapore.
\bibitem{c10}Mark W. Spong, Seth Hutchinson, and M. Vidyasagar. Robot Dynamics and Control
\end{thebibliography}
\end{document}